\documentclass[conference]{IEEEtran}
\IEEEoverridecommandlockouts
\usepackage{cite}
\usepackage{amsmath,amssymb,amsfonts}
\usepackage{algorithmic}
\usepackage{graphicx}
\usepackage{textcomp}
\usepackage{stfloats}
\usepackage{url}
\usepackage{marvosym}
\usepackage{verbatim}
\usepackage[table]{xcolor}
\usepackage{xcolor}
\usepackage{booktabs}
\usepackage{multirow}
\usepackage[pagebackref=true,breaklinks=true,colorlinks,bookmarks=false]{hyperref}

\definecolor{mycolor}{RGB}{198,234,251}
\definecolor{unitcolor}{RGB}{235,235,235}

\definecolor{mycolor1}{RGB}{62,128,128}
\definecolor{mycolor2}{RGB}{30,18,0}
\definecolor{mycolor3}{RGB}{255,230,230}

\newcommand{\colorhat}[1]{{\color{red}\hat{\color{black}{#1}}}}
\newcommand{\spthickhline}{%
    \noalign {\ifnum 0=`}\fi \hrule height 1.5pt
    \futurelet \reserved@a \@xhline
}
\def\BibTeX{{\rm B\kern-.05em{\sc i\kern-.025em b}\kern-.08em
    T\kern-.1667em\lower.7ex\hbox{E}\kern-.125emX}}

\makeatletter
\newcommand{\linebreakand}{%
  \end{@IEEEauthorhalign}
  \hfill\mbox{}\par
  \mbox{}\hfill\begin{@IEEEauthorhalign}
}
\makeatother
    
\begin{document}

\title{Uncertainty-Participation  Context Consistency Learning for Semi-supervised Semantic Segmentation
\thanks{\hspace{-1.2em}
 \textsuperscript{(\Letter)}Corresponding author: Yanhui Gu. This work is supported by the Natural Science Foundation
of China (Nos. 62377029, 92370127 and 22033002).}
}

% \author{\IEEEauthorblockN{Jianjian Yin}
% \IEEEauthorblockA{\textit{School of Artificial Intelligence} \\
% \textit{Nanjing Normal University}\\
% Nanjing, China \\
% JianJYin@gmail.com}
% \and
% \and
% \and
% \hspace{2cm}\IEEEauthorblockN{4\textsuperscript{th} Given Name Surname}
% \and
% % \and
% % \IEEEauthorblockN{6\textsuperscript{th} Given Name Surname}
% % \IEEEauthorblockA{\textit{dept. name of organization (of Aff.)} \\
% % \textit{name of organization (of Aff.)}\\
% % City, Country \\
% % email address or ORCID}
% }

\author{\IEEEauthorblockN{Jianjian Yin}
\IEEEauthorblockA{\textit{School of Artificial Intelligence} \\
\textit{Nanjing Normal University}\\
Nanjing, China \\
JianJYin@gmail.com}
\and
\IEEEauthorblockN{Yi Chen}
\IEEEauthorblockA{\textit{School of Artificial Intelligence} \\
\textit{Nanjing Normal University}\\
Nanjing, China \\
cs\_chenyi@njnu.edu.cn}
\and
\IEEEauthorblockN{Zhichao Zheng}
\IEEEauthorblockA{\textit{School of Artificial Intelligence}  \\
\textit{Nanjing Normal University}\\
Nanjing, China \\
zheng\_zhichaoX@163.com}
%\and
\linebreakand 
\IEEEauthorblockN{Junsheng Zhou}
\IEEEauthorblockA{\textit{School of Artificial Intelligence}  \\
\textit{Nanjing Normal University}\\
Nanjing, China \\
zhoujs@njnu.edu.cn}
\and
\IEEEauthorblockN{Yanhui Gu \textsuperscript{(\Letter)}}
\IEEEauthorblockA{\textit{School of Artificial Intelligence}  \\
\textit{Nanjing Normal University}\\
Nanjing, China \\
gu@njnu.edu.cn
}
}

\maketitle
\begin{abstract}
Semi-supervised semantic segmentation has attracted considerable attention for its ability to mitigate the reliance on extensive labeled data. However, existing consistency regularization methods only utilize high certain pixels with prediction confidence surpassing a fixed threshold for training, failing to fully leverage the potential supervisory information within the network.  Therefore, this paper proposes the \textbf{U}ncertainty-participation  \textbf{C}ontext \textbf{C}onsistency \textbf{L}earning (\textbf{UCCL}) method to explore richer supervisory signals. Specifically, we first design the semantic backpropagation update (SBU) strategy to fully exploit the knowledge from uncertain pixel regions, enabling the model to learn consistent pixel-level semantic information from those areas. Furthermore, we propose the class-aware knowledge regulation (CKR) module to facilitate the regulation of class-level semantic features across different augmented views, promoting consistent learning of class-level semantic information within the encoder.
Experimental results on two public benchmarks demonstrate that our proposed method achieves state-of-the-art performance. Our code is available at \url{https://github.com/YUKEKEJAN/UCCL}.
\end{abstract}

\begin{IEEEkeywords}
Semi-supervised semantic segmentation, Uncertainty-participation  context consistency learning, Semantic backpropagation update, Class-aware knowledge regulation.
\end{IEEEkeywords}

\begin{figure}[t]
\centering
\includegraphics[width=1\linewidth]{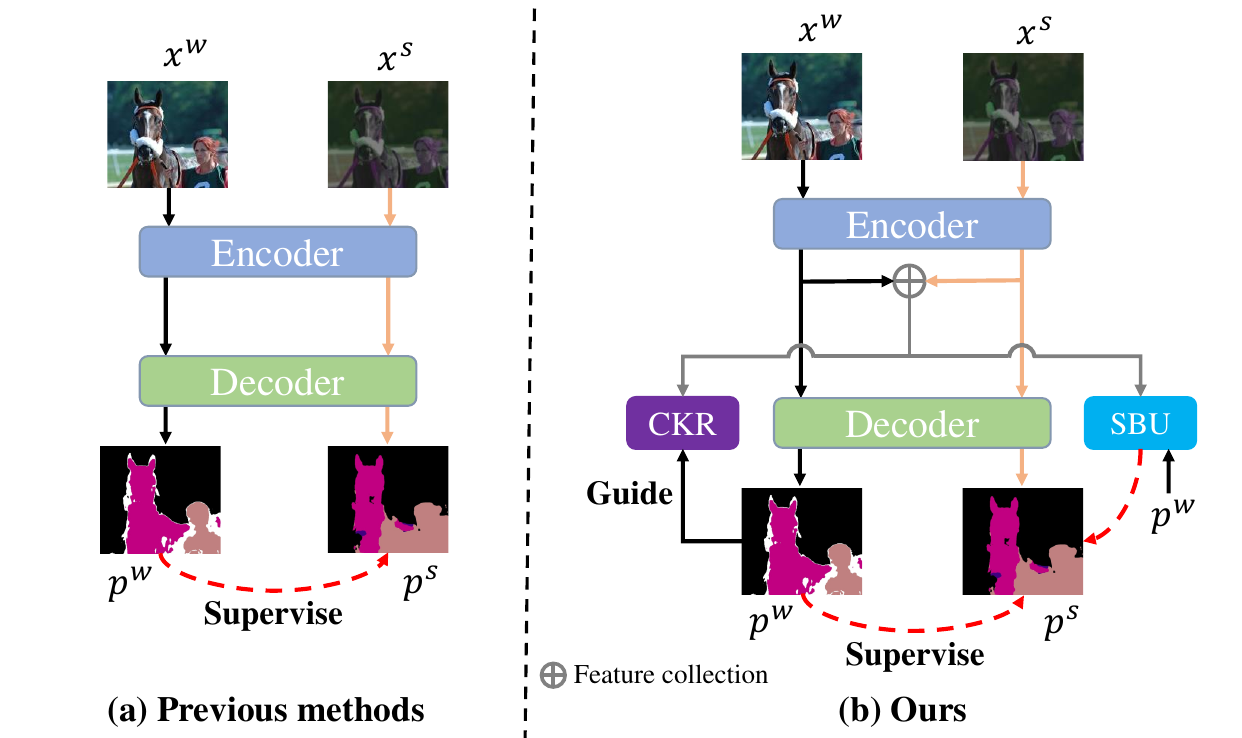}
% \vspace{-0.1cm}
\caption{Comparison with previous methods. (a) The existing methods exclusively leverage labels of high certain pixels (non-white regions in $p^w$) for supervised training to ensure consistency in  prediction results ($p^s$ and $p^w$). (b) Our approach focuses more on utilizing information from  uncertain pixels  (white regions in $p^w$) in strongly and weakly augmented views ($x^s$ and $x^w$) by semantic backpropagation update (SBU), while also exploring potential consistent semantic information via class-aware knowledge regulation (CKR).}
\vspace{-0.2cm}
\label{fig1}
\end{figure}

\section{Introduction}
\label{sec:intro}
Although deep learning-based semantic segmentation \cite{wu2024image,chen2023generative,zhang2023delivering,yin2024Class,yinCLASS,yang2022cross,chen2023multi,li2022deep,zhou2022rethinking,jin2024idrnet,chen2022saliency,chen2025knowledge,wu2025adaptive,wu2023top} for pixel-level classification tasks has made significant progress in recent years,  the performance of current methods \cite{wu2023sparsely,song2022fully,guo2022segnext,xu2023pidnet,sung2024contextrast} heavily depends on the quantity and quality of labeled data, which are time-consuming and labor-intensive to obtain. 
Consequently, semi-supervised semantic segmentation (\textbf{S4})  \cite{wang2022semi,kwon2022semi,chen2021semi,yuan2023,yin2024Class,ouali2020semi,yang2022st++,yin2023semi}, which train models using limited  labeled data along with substantial unlabeled data, have been gaining increasing research interest. A mainstream direction in the domain of \textbf{S4} is consistency regularization \cite{xie2020unsupervised}, with the goal of encouraging the network to produce equivalent outputs for different augmented views of the same unlabeled data. Some methods \cite{liu2022perturbed,yuan2023,wang2023conflict} introduce  various perturbations into the network to achieve  prediction consistency. Additionally, other methods \cite{yang2023revisiting,fan2023conservative,zhao2023augmentation} concentrate on designing  diverse data augmentation strategies to improve the  generalization of the model. While the aforementioned methods exhibit  promising experimental performance, a pivotal question remains: \textit{the insufficient utilization of potential supervisory information within the network}.

\begin{figure*}[t]
\centering
\includegraphics[width=1\linewidth]{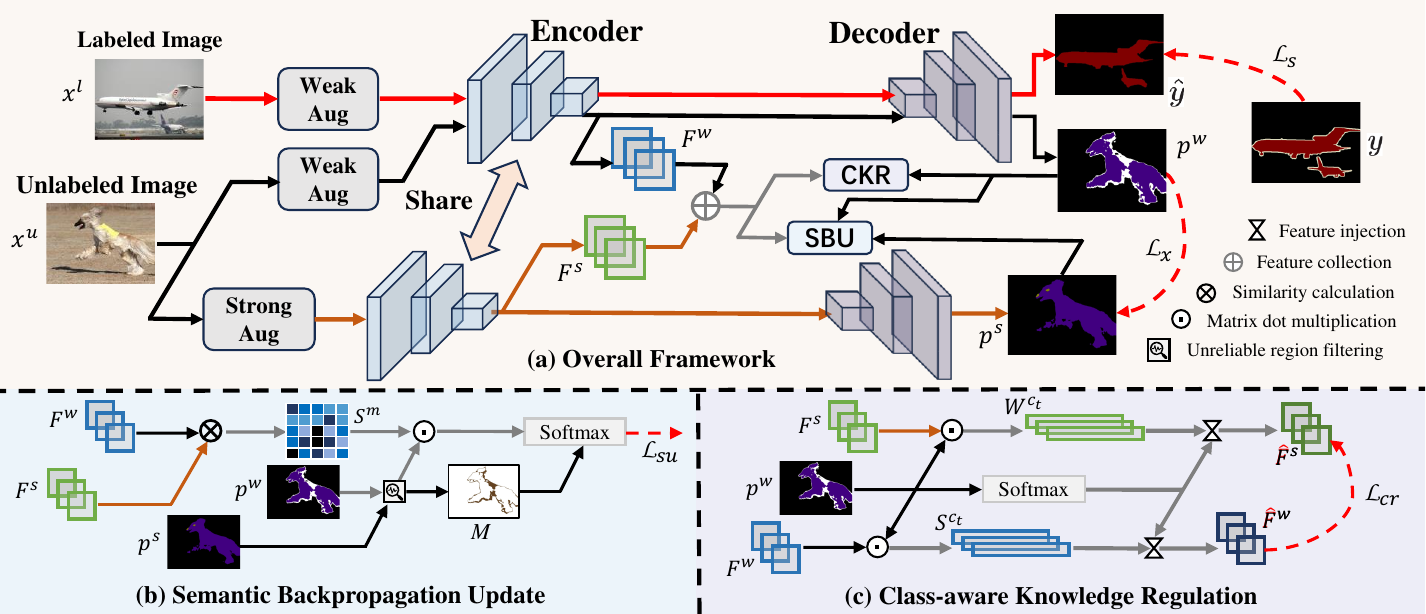}
% \vspace{-0.5cm}
\caption{Illustration of our UCCL method. The paired encoder and decoder parameters are shared.}
\label{fig2}
\vspace{-0.2cm}
\end{figure*}

As shown in Fig. \ref{fig1}, existing methods concentrate on using labels of high certainty pixels from weakly augmented view prediction result ($p^w$), where confidence exceeds a fixed threshold, to supervise predictions ($p^s$) in strongly augmented view, without considering the potential of  uncertain pixel information. 
Moreover, intuitively, the process by which a high-performance network produces consistent prediction results for different augmented views ($x^s$ and $x^w$) can be divided into two steps: First, the encoder generates highly consistent contextual (semantic) features for $x^s$ and $x^w$. Second, the decoder produces consistent prediction results based on the contextual features. This inherent semantic consistency can be leveraged to improve the robustness of the encoder. However, current methods \cite{yang2023revisiting,yuan2023,zhao2023augmentation} neglect these aspects, thereby restricting further advancements in the \textbf{S4}.

To alleviate the above issue, this paper proposes the \textbf{U}ncertainty-participation  \textbf{C}ontextual \textbf{C}onsistency \textbf{L}earning (\textbf{UCCL}) method, aiming to explore the potential of uncertain pixel regions and the contextual consistency across different augmented views, enabling the full utilization of potential supervisory information. The semantic backpropagation update (SBU) strategy, proposed by us, weights  the supervisory loss in a pixel-specific way, it does so by exploiting the pairwise similarity semantic information between uncertain region pixels in different augmented views, thus inducing the network to learn consistent pixel-level semantic information from these uncertain regions. Additionally, class-level semantic information is  utilized by our designed class-aware knowledge regulation (CKR) module, which collaborates with the SBU module to further enhance the robustness of the encoder.  Our contributions can be summarized as follows: (1) We propose the \textbf{UCCL} method to fully utilize the potential supervisory information within the network. (2) The semantic backpropagation update strategy is proposed  to extract pixel-level semantic consistency information from uncertain pixel regions. (3) We design the class-aware knowledge regulation module to learn consistent class-level semantic information across different augmented views. Extensive experiments on the Pascal VOC2012 and  Cityscapes  datasets demonstrate that our method has achieved state-of-the-art performance.

% \vspace{-0.2cm}
\section{METHODOLOGY}
%\vspace{-0.15cm}
\subsection{Overall Framework}
% \vspace{-0.1cm}
The goal of semi-supervised semantic segmentation is to explore more classification information learned from massive unlabeled data ${S^u} = \{ ({x_i^u})\} _{i = 1}^{|{S^u}|}$, based on the knowledge acquired from a small amount of labeled data ${S^l} = \{ ({x_i^l},{y_i})\} _{i = 1}^{|{S^l}|}$. Labeled data $x^l$ is initially processed with weak augmentation  before being fed into the network for prediction $\hat y$. The prediction is then supervised with the corresponding ground-truth $y$, which directs the learning process on the labeled data:
\begin{equation}
\vspace{-0.1cm}
    \label{eq1}
    {\mathcal{L}_s} = \frac{1}{{{B_l}}}\sum\limits_{i = 1}^{{B_l}} {\sum\limits_{j = 1}^{H{\rm{ \times }}W} {{\ell _{ce}}({{\hat y}_{(i,j)}},{y_{(i,j)}})} },
    \vspace{-0.1cm}
\end{equation}
$\mathcal{L}_s$, $B_l$, and $\ell _{ce}(\cdot)$ represent the labeled supervision loss, batch size, and cross-entropy loss function, respectively. $H$ and $W$ denote the size of the image. $y_{(i,j)}$ is the label of the $j$-th pixel in the $i$-th image. Unlabeled data $x^u$ undergoes strong and weak augmentations, producing a strongly augmented view $x^s$ and a weakly augmented view $x^w$, respectively. These views are subsequently fed into the encoder $\Psi(\cdot)$ to generate the corresponding strongly and weakly augmented view feature representations ($F^s$ and $F^w$):
\begin{equation}
\vspace{-0.1cm}
    \label{eq2}
    {F^s} = \Psi ({x^s}); \quad {F^w} = \Psi ({x^w}).
    % \vspace{-0.1cm}
\end{equation}
Next, $F^w$ is fed into the decoder $\Theta(\cdot)$ to obtain the prediction result $p^w$ for the weakly augmented view, and similarly for prediction result $p^s$ for the strongly augmented view:
\begin{equation}
\vspace{-0.1cm}
    \label{eq3}
    {p^s} = \Theta ({F^s});\quad {p^w} = \Theta ({F^w}).
    % \vspace{-0.1cm}
\end{equation}
Following previous methods \cite{yuan2023,yang2023revisiting}, we also use the labels of high certainty regions in $p^w$ to supervise $p^s$, generating an unsupervised loss $\mathcal{L}_x$ that facilitates consistency regularization:
\begin{equation}
\vspace{-0.15cm}
    \label{eq4}
     {\mathcal{L}_x} = \frac{1}{{{B_u}}}\sum\limits_{i = 1}^{{B_u}} {\sum\limits_{j = 1}^{H{\rm{ \times }}W} {(\varphi (p_{(i,j)}^w) > \tau )\cdot{\ell _{ce}}(p_{(i,j)}^s,p_{(i,j)}^w}}).
    %\vspace{-0.1cm}
\end{equation}
$\varphi(\cdot)$ is used to select the maximum probability value, and $\tau$ refers to the predefined threshold.   The \textit{Semantic Backpropagation Update} section will analyze the use of information from uncertain pixel regions, while the \textit{Class-aware Knowledge Regulation} section will explore the inherent semantic consistency  across different augmented views.

% \textcolor{red}{The semantic backpropagation update (SBU) strategy computes the weight of the supervision loss for uncertain pixel regions by assessing  the pairwise similarity between the regions in $F^s$ and $F^w$, guiding the network to  learn consistent pixel-level semantic information. The class-aware knowledge regulation (CKR) module encourages the network to learn consistent class-level semantic information from different augmented images.}

\subsection{Semantic Backpropagation Update}
\label{SBU}
Existing methods \cite{yang2023revisiting,yuan2023} do not take into account the potential of uncertain pixels when calculating supervision loss $\mathcal{L}_x$, which motivated the design of the semantic backpropagation update (SBU) strategy.

We use Eq. \ref{eq2} to obtain the feature representations of the strongly and weakly augmented views. Subsequently, we calculate the pairwise similarity matrix $S^m$ between features of different augmented views and the uncertainty pixel region mask $M$: 
\vspace{-0.15cm}
\begin{equation}
    \label{eq5}
     {S^m} = \Omega ({F^s},{F^w}),\quad {M} = \xi (({p^w}) < \tau ).
     \vspace{-0.05cm}
\end{equation}
$\Omega(\cdot)$ is employed to calculate the pairwise similarity between features. $\xi(\cdot)$ performs conditional judgment and then downsamples the size to match that of $F^w$.  For class $c_t$ in the uncertain pixel regions of the image, we extract the set of pixel similarities $V^{c_t}$ at the corresponding locations based on the pairwise similarity matrix $S^m$:
\vspace{-0.05cm}
\begin{equation}
    \label{eq6}
    {V^{{c_t}}} = \{ \xi(\arg \max ({p^w}) = {c_t})\; \odot  {M} \odot  {S^m}\}.
    \vspace{-0.05cm}
\end{equation}
$\odot$ is the matrix dot multiplication. Finally, we apply the softmax function $\delta(\cdot)$ to the values in $V^{{c_t}}$ to obtain weights for the supervision loss of corresponding pixels in the uncertain regions, which are then used to construct the semantic backpropagation update loss $\mathcal{L}_{su}$:
\vspace{-0.09cm}
\begin{equation}
    \label{eq7}
    {{\cal L}_{su}} = \frac{1}{{{B_u}}}\sum\limits_{i = 1}^{{B_u}} {\sum\limits_{t = 1}^C {\delta ( - V_i^{{c_t}})}  \odot  M_i \odot  {\ell _{ce}}(p_i^s,p_i^w)}.
    \vspace{-0.01cm}
\end{equation}
$C$ denotes the number of classes in the dataset. Eq. \ref{eq7} can be understood as follows: the lower the pairwise similarity of uncertain pixels within the same class across different augmented views, the higher the corresponding loss weight, while higher similarity results in lower weight. The semantic backpropagation update strategy encourages the network to focus on training uncertain regions in a pixel-specific manner, dynamically updating the loss weight during subsequent training to learn consistent pixel-level semantic information.

\subsection{Class-aware Knowledge Regulation}

Compared to pixel-level semantic information, class-level semantic information can also assist the model in contextual semantic perception. Unlike the SBU strategy, which concentrates solely on the pixel information of uncertain regions, the class-aware knowledge regulation module utilizes information from all pixels in the image to enforce semantic information consistency at the class level  across different augmented views. 

We first use the prediction results $p^w$ to filter the class $c_t$ features in both $F^w$ and $F^s$:
\begin{equation}
% \vspace{-0.2cm}
    \label{eq8}
    \begin{array}{l}
{S^{c_t}} = \left\{ {\xi (argmax({p^w}) = {c_t}) \odot {F^s}} \right\};\\
{W^{c_t}} = \left\{ {\xi (argmax({p^w}) = {c_t}) \odot {F^w}} \right\}.
\end{array}
\end{equation}
$S^{c_{t}}$ and $W^{c_{t}}$ correspond to the sets of features for class 
$c_t$. 
Next, we obtain the probability value  for each pixel of class $c_t$ in $p^w$, and perform softmax operation $\delta(\cdot)$ to get the corresponding weight set $H^{c_t}$.
The class-level semantic feature $R^{c_t}$ for category $c_t$ in the current weakly augmented view are obtained using the following formula:
\begin{equation}
% \vspace{-0.1cm}
    \label{eq9}
R^{c_{t}}=\sum_{k=1}^{k=E}H_{k}^{c_t}\cdot W_{k}^{c_{t}},  \quad D^{c_{t}}=\sum_{k=1}^{k=E}H_{k}^{c_t}\cdot S_{k}^{c_{t}}.
    % \vspace{-0.1cm}
\end{equation}
Similarly, $D^{c_{t}}$ represents the class-level feature of category $c_t$  for the corresponding strongly augmented view. $E$ is the total number of pixels for class $c_t$.
The class-level semantic features of each  class $\left \{ c_{z}  \right \}_{z=1}^{C} $ are also obtained using Eqs. \ref{eq8} and \ref{eq9}.
Next, we construct the class-level semantic feature representations ($\colorhat{F}^w$ and $\colorhat{F}^s$) based on the prediction results $p^w$ and the class-level features for each class:
\begin{equation}
\vspace{-0.05cm}
    \label{eq10}
    \colorhat{F}_{(i,j)}^s = {D^{{c_z}}},\;\;\colorhat{F}_{(i,j)}^w = {R^{{c_z}}}\;\;if\;\arg \max (p_{(i,j)}^w = {c_z}).\;
    % \vspace{-0.1cm}
\end{equation}
 Finally, we compute the class-aware knowledge regulation loss ${\cal L}_{cr}$ as given by the formula below:
\begin{equation}
\vspace{-0.1cm}
    \label{eq11}
    {{\cal L}_{cr}} = 1 - \frac{1}{{{B_u}}}\frac{1}{N}\sum\limits_{i = 1}^{{B_u}} {\sum\limits_{j = 1}^N {\Omega (\colorhat{F}_{(i,j)}^s),\colorhat{F}_{(i,j)}^w})}.
    % \vspace{-0.1cm}
\end{equation}
$N$ is the number of features.  The class-aware knowledge regulation module uses the class-level semantic feature representations of weakly augmented views to regulate those of strongly augmented views, allowing the model to learn potential class-level consistent semantic information.

The total loss function ${\cal L}$ used for network training is as follows:
\begin{equation}
\vspace{-0.1cm}
    \label{eq12}
    {\cal L} = {{\cal L}_s} + {{\cal L}_x} + \alpha {{\cal L}_{su}} + \beta {{\cal L}_{cr}}.
\end{equation}
$\alpha$ and $\beta$ are empirically set to 0.015 and 0.02, respectively.

\section{EXPERIMENT}
\label{experiment}
\subsection{Experiment Setup}
\textcolor{white}{whl}\textbf{Datasets.} We evaluate  the performance of our method on two standard semantic segmentation datasets: Pascal VOC2012 \cite{everingham2015pascal} and Cityscapes \cite{cordts2016cityscapes}. The Pascal VOC2012 dataset comprises 21 classes, with 1,464 precisely annotated images for training and 1,449 images for validation. Subsequently, it was augmented by the SBD \cite{hariharan2011semantic} dataset to include 10,582 images. For model training, labeled data is selected from the 1,464  high-quality images, with the remaining images used as unlabeled data. Therefore, \textit{the Full (1464) setting in Table \ref{tab_1} includes 1,464 labeled  and 9,118 unlabeled images.} The Cityscapes dataset contains 19 classes, comprising 2,975 images for training and 500 images for validation.

\textbf{Implementation Details.} Our model is implemented using the Pytorch framework. For the Pascal VOC2012 / Cityscapes datasets, the initial learning rates are set to 0.001 / 0.005,  and the number of training epochs is 80 / 240, respectively. The input image sizes are 321$\times$321 / 801$\times$801, with a batch size of 8 for all experiments. We use ResNet-50 \cite{he2016deep} as the encoder and DeepLabv3+ \cite{chen2018encoder} as the decoder for all experiments. All experiments are conducted on four NVIDIA RTX A6000 GPUs. 
Weak augmentation includes horizontal and vertical flipping, while strong augmentation not only incorporates the operations of weak augmentation but also includes color jittering, graying, and Gaussian blur.
Following other state-of-the-art methods (UniMatch \cite{yang2023revisiting} and MKD \cite{yuan2023}), we use mean Intersection over Union (mIoU) as the evaluation metric. 

\begin{table*}[t]
    \centering
    \renewcommand\arraystretch{1}
    \begin{minipage}{0.45\textwidth}
        \centering
         \caption{Comparison of the performance with other  methods on the Pascal VOC2012 dataset. 1/n represents the ratio of labeled data, with the rest being unlabeled. $\dag$ represents the results reproduced under the same settings.}
         \vspace{-0.2cm}
        \setlength{\tabcolsep}{0.85mm}{
        \begin{tabular}{c|cccc}
            \toprule
            {\footnotesize \textbf{Methods}} & {\footnotesize \textbf{1/8 (183)}} & {\footnotesize \textbf{1/4 (366)}} & {\footnotesize \textbf{ 1/2 (732)}} & {\footnotesize \textbf{Full (1464)}}  \\
            \midrule
            {\footnotesize SupOnly} & {\footnotesize 52.26} &   {\footnotesize 61.65} &   {\footnotesize 66.72}  &  {\footnotesize 72.94} \\
            \midrule
            {\footnotesize CPS \cite{chen2021semi}} {\tiny \textcolor{gray}{[CVPR 21]}} & {\footnotesize 67.42}&  {\footnotesize 71.71} &  {\footnotesize 75.88} &  -  \\
            % {\footnotesize PC$^2$Seg \cite{zhong2021pixel}} {\tiny \textcolor{gray}{[ICCV 21]}} & {\footnotesize 64.63}&  {\footnotesize 67.62} &  {\footnotesize 70.90} &  {\footnotesize 72.26}  \\
            {\footnotesize ELN \cite{kwon2022semi}} {\tiny \textcolor{gray}{[CVPR 22]}}  & {\footnotesize 73.20} &  {\footnotesize 74.63} & - & - \\
            {\footnotesize CPCL \cite{fan2023conservative}}  {\tiny \textcolor{gray}{[TIP 23]}} & {\footnotesize 67.02} & {\footnotesize 72.14} & {\footnotesize 74.25} & - \\
            {\footnotesize MKD \cite{yuan2023}}  {\tiny \textcolor{gray}{[ACMMM 23]}} & {\footnotesize 66.74} &  {\footnotesize 71.01} &  {\footnotesize 72.73}  &  {\footnotesize 78.14}  \\
            % {\footnotesize AugSeg \cite{zhao2023augmentation}} {\tiny \textcolor{gray}{[CVPR 23]}} & {\footnotesize 72.17}&  {\footnotesize 76.17} &  {\footnotesize 77.40} &  {\footnotesize 78.82} \\
            {\footnotesize UniMatch \cite{yang2023revisiting}} {\tiny \textcolor{gray}{[CVPR 23]}}  & {\footnotesize 72.48}  &  {\footnotesize 75.96}  &  {\footnotesize 77.39} &  {\footnotesize 78.70} \\
            {\footnotesize ESL \cite{Ma_2023_ICCV}} {\tiny \textcolor{gray}{[ICCV 23]}} & {\footnotesize 69.50} &  {\footnotesize 72.63}  &   {\footnotesize 74.69}  &   {\footnotesize 77.11} \\
            {\footnotesize RankMatch$\dag$\cite{mai2024rankmatch}} {\tiny \textcolor{gray}{[CVPR 24]}} & {\footnotesize 73.11} &  {\footnotesize 76.33}  &   {\footnotesize 77.51}  &   {\footnotesize 78.99} \\
            \midrule
            \rowcolor{mycolor3} \textbf{{\footnotesize Ours}}  & \textbf{{\footnotesize 74.09}} & \textbf{{\footnotesize 77.08}}  & \textbf{{\footnotesize 78.18}} &	\textbf{{\footnotesize 79.43}} \\
            \bottomrule
        \end{tabular}}
        \label{tab_1}
    \end{minipage}
    \hspace{0.05\textwidth}
    \begin{minipage}{0.45\textwidth}
        \centering
        \caption{Comparison of the performance with other  methods on the Cityscapes dataset. 1/n represents the ratio of labeled data, with the rest being unlabeled. $\dag$ represents the results reproduced under the same settings.}
        \vspace{-0.2cm}
        \setlength{\tabcolsep}{0.85mm}{
        \begin{tabular}{c|cccc}
            \toprule
            {\footnotesize \textbf{Methods}} & {\footnotesize \textbf{1/16 (186)}} & {\footnotesize \textbf{1/8 (372)}} & {\footnotesize \textbf{1/4 (744)}} & {\footnotesize \textbf{1/2 (1488)}}  \\
             \midrule
            {\footnotesize SupOnly} & {\footnotesize 63.30} & {\footnotesize 70.20} & {\footnotesize 73.10} & {\footnotesize 76.60} \\
            \midrule
            {\footnotesize CPS \cite{chen2021semi}}  {\tiny \textcolor{gray}{[CVPR 21]}} & {\footnotesize 69.79} & {\footnotesize  74.39} & {\footnotesize 76.85} & {\footnotesize 78.64} \\
            % {\footnotesize PS-MT \cite{liu2022perturbed}} {\tiny \textcolor{gray}{[CVPR 22]}} & - & {\footnotesize 75.76} &  {\footnotesize 76.92} & {\footnotesize 77.64} \\
            {\footnotesize ELN \cite{kwon2022semi}} {\tiny \textcolor{gray}{[CVPR 22]}} & - & {\footnotesize 70.33} &  {\footnotesize 73.52} & {\footnotesize 75.33} \\
            % {\footnotesize U2PL \cite{wang2022semi}} {\tiny \textcolor{gray}{[CVPR 22]}} & {\footnotesize 69.00} & {\footnotesize 73.00} &  {\footnotesize 76.30} & {\footnotesize 77.10} \\
            {\footnotesize CPCL \cite{fan2023conservative}}  {\tiny \textcolor{gray}{[TIP 23]}} & {\footnotesize 69.92} & {\footnotesize 74.60} & {\footnotesize 76.98} & {\footnotesize 78.17}\\
            {\footnotesize CCVC \cite{wang2023conflict}} {\tiny \textcolor{gray}{[CVPR 23]}} & {\footnotesize 74.90} & {\footnotesize 76.40} & {\footnotesize 77.30} & - \\
            {\footnotesize UniMatch \cite{yang2023revisiting}} {\tiny \textcolor{gray}{[CVPR 23]}}  & {\footnotesize 75.03}  &  {\footnotesize 76.77}  &  {\footnotesize 77.49} &  {\footnotesize 78.60} \\
             {\footnotesize ESL \cite{Ma_2023_ICCV}} {\tiny \textcolor{gray}{[ICCV 23]}} & {\footnotesize 71.07} &  {\footnotesize 76.25}  &   {\footnotesize 77.58}  &   {\footnotesize 78.92} \\
             {\footnotesize CorrMatch$\dag$\cite{Sun_2024_CVPR}} {\tiny \textcolor{gray}{[CVPR 24]}} & {\footnotesize 75.29} &  {\footnotesize 76.85}  &   {\footnotesize 77.91}  &   {\footnotesize 78.65} \\
             \midrule
            \rowcolor{mycolor3} \textbf{{\footnotesize Ours}}  & \textbf{{\footnotesize 75.76}} & \textbf{{\footnotesize 77.21}}  & \textbf{{\footnotesize 78.22}} &	\textbf{{\footnotesize 79.54}} \\
            \bottomrule
        \end{tabular}}
        \label{tab_2}
    \end{minipage}
   \vspace{-0.4cm}
\end{table*}

\begin{table}[t]
\caption{Component analysis of semantic backpropagation update (SBU)  and class-aware knowledge regulation (CKR) under the 1/2 (732)  setting on the Pascal VOC2012 dataset. "Time" shows the duration to process a batch of data under the same setting.}
\renewcommand\arraystretch{1}
\vspace{-0.2cm}
\centering
        \begin{tabular}{ccc|ccc}
           \toprule
        {\footnotesize Baseline} & {\footnotesize SBU}  & {\footnotesize CKR} & {\footnotesize mIoU $\uparrow$} &  {\footnotesize Params $\downarrow$} & {\footnotesize Time $\downarrow$}  \\
        \midrule
        {\footnotesize $\checkmark$} & & & {\footnotesize 77.39} & {\footnotesize 40.48M} & {\footnotesize 0.199s}\\
        {\footnotesize $\checkmark$} & {\footnotesize $\checkmark$} & & {\footnotesize 77.80} & {\footnotesize 40.48M} & {\footnotesize 0.207s} \\
        {\footnotesize $\checkmark$} &  & {\footnotesize $\checkmark$} & {\footnotesize 77.84} & {\footnotesize 40.48M} & {\footnotesize 0.234s} \\
        {\footnotesize $\checkmark$} & {\footnotesize $\checkmark$} & {\footnotesize $\checkmark$} & {\footnotesize \textbf{78.18}} & {\footnotesize 40.48M} & {\footnotesize 0.257s}\\
        \bottomrule
        \end{tabular}
    \label{tab3}
    \vspace{-0.241cm}
\end{table}

\subsection{Comparison with State-of-the-Art Methods}
\begin{figure}[t]
\centering
\includegraphics[width=1\linewidth]{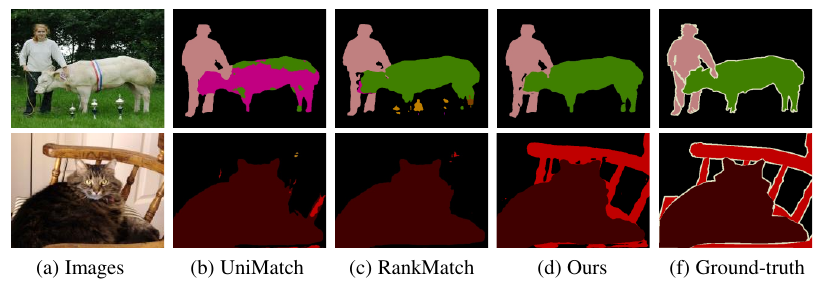}
\vspace{-0.6cm}
\caption{Comparison of visualization results with  state-of-the-art methods under the Full (1464)  setting on  the Pascal VOC2012 dataset.}
\vspace{-0.237cm}
\label{fig3}
\end{figure}

Tables \ref{tab_1} and \ref{tab_2} show the comparison of experimental performance between our proposed UCCL method and other state-of-the-art methods on different datasets. 'SupOnly' means training only with labeled data, without using unlabeled data.  For the Pascal VOC2012 dataset, our method outperforms RankMatch \cite{mai2024rankmatch} by 0.98\%, 0.75\%, 0.67\%, and 0.44\% mIoU in the 1/8, 1/4, 1/2, and Full settings, respectively. Similarly, the results also highlight the superiority of our method compared to ESL \cite{Ma_2023_ICCV}. For the Cityscapes dataset, our method achieves mIoU improvements of 4.69\%, 0.96\%, 0.64\%, and 0.62\% over ESL \cite{Ma_2023_ICCV} in the 1/16, 1/8, 1/4, and 1/2 settings, respectively.
Additionally, the segmentation results in Fig. \ref{fig3} demonstrate that our method places a greater emphasis on context consistency information. The state-of-the-art performance across all settings on Pascal VOC2012 and Cityscapes datasets  is attributed to the effective  utilization of potential supervisory information within the network.

\subsection{Ablation Studies}
We conduct detailed ablation experiments on semantic backpropagation update (SBU) strategy and class-aware knowledge regulation (CKR)  module, as shown in Table \ref{tab3}. We use UniMatch \cite{yang2023revisiting} as the baseline. 
When used alone, the SBU strategy improves the performance of the model by 0.41\% mIoU, highlighting the benefits of fully leveraging the pixel-level semantic consistency information from uncertain regions. Integrating the CKR module into the model alone results in a 0.45\% mIoU performance improvement, which demonstrates the effectiveness of class-level semantic consistency information. Simultaneously incorporating SBU and CKR improves the performance of the model by 0.79\% mIoU without introducing significant computational overhead, indicating that SBU and CKR can complement each other and facilitate robust  context consistency learning of the model.

\section{Conclusion}
This paper proposes the \textbf{U}ncertainty-participation  \textbf{C}ontextual \textbf{C}onsistency \textbf{L}earning (\textbf{UCCL}) method, which fully leverages potential supervisory information to facilitate contextual consistency learning within the network. Initially, we design the semantic backpropagation update strategy to explore  consistent  pixel-level semantic information in uncertain regions. Furthermore, we propose the class-aware knowledge regulation module to extract class-level consistent semantic information across different augmented views. Extensive comparative experiments on two datasets show that our method achieves state-of-the-art performance.

\bibliographystyle{IEEEbib}
\bibliography{refs}

\end{document}